# COLOR AND GRADIENT FEATURES FOR TEXT SEGMENTATION FROM VIDEO FRAMES

[a]P. Shivakumara, [b]D. S. Guru, and [b]H.T. Basavaraju

[a]Department of Computer Science, School of Computing, National University of Singapore, Singapore.

[b]Department of Studies in Computer Science, University of Mysore, Mysore, Karnataka, India.

[a]hudempsk@yahoo.com, [b]dsg@compsci.uni_mysore.ac.in , [b]basavaraju.com@gmail.com

**Abstract.** Text segmentation in a video is drawing attention of researchers in the field of image processing, pattern recognition and document image analysis because it helps in annotating and labeling video events accurately. We propose a novel idea of generating an enhanced frame from the R, G, and B channels of an input frame by grouping high and low values using Min-Max clustering criteria. We also perform sliding window on enhanced frame to group high and low values from the neighboring pixel values to further enhance the frame. Subsequently, we use k-means with k=2 clustering algorithm to separate text and non-text regions. The fully connected components will be identified in the skeleton of the frame obtained by k-means clustering. Concept of connected component analysis based on gradient feature has been adapted for the purpose of symmetry verification. The components which satisfy symmetric verification are selected to be the representatives of text regions and they are permitted to grow to cover their respective region fully containing text. The method is tested on variety of video frames to evaluate the performance of the method in terms of recall, precision and f-measure. The results show that method is promising and encouraging.

## 1 Introduction

The retrieval of text information from images and videos is a very hot research area and it has gained increasing attention in the recent years. Text in images and video sequences can provide very useful semantic information which would be a good key to describe the image content and help a machine to understand it. This would help in bridging the gap between the low level and high level features [1-3]. Therefore, text segmentation and localization in video plays a vital role in several applications such as indexing video retrieval, video event identification, video understanding and video tracking etc. However, text segmentation is still a challenging problem because of the low resolution, complex background, unconstrained colors, sizes, and alignments of the characters [4-5].

The methods [5-7] in document analysis cannot be used directly for video text segmentation as these methods require complete shape of the character components and high resolution images with plain background. These constraints may not be true for video as video usually have low resolution and complex background images. Therefore, the document analysis methods are not suitable. It is also observed from the literature on text detection from natural scene images that the methods [8-10] one or other ways extract features based on shape of the characters. Though these methods work for complex background images but they require high resolution images. Therefore, the scene text detection from natural scene methods cannot be applied on directly on video images.

In general, the existing methods on video text detection can be classified as connected component based [11-12], texture based [13-15] and edge and gradient based [16-19] methods. The connected component based methods are same as documents analysis methods. Therefore, they work only for high resolution and simple background images. Texture based methods work well for complex background images but they require expensive classifier to classify text and non-text components. In addition, these methods are sensitive to font, font size and multi-script. To overcome the problems of the above categories, the edge and gradient based methods are proposed. These methods are fast compared to the above two categories. Recently, Eigen based method [20] is proposed to detect text in video where the method explores Eigen value analysis to classify text and non-text pixels for the both low and high resolution images. Since the method involves Eigen value analysis and gradient information, it is said to be expensive and give low precision for complex background images. In the same way, the method [21] based on run-lengths between inter and intra text component is proposed for video text detection. This method said to be simple and effective for both low resolution and high resolution video images. However, the methods produce more false positives due to text like edges in the background and hence the methods report high false positive rate and low precision. It is noted from the literature review that none of the existing methods give perfect solution to video text segmentation. In addition, none of the methods explore the combination of color values for text enhancement and symmetry based on stroke width for finding text representatives.

Hence, in this paper, we propose a novel method to explore the color values and symmetry concepts to accurate video text segmentation. The main contribution of the paper is that sharpening text information and widening gap between text and non-text using color values mapping and proposing new symmetry to identify the text representatives, which eliminates almost all background information since the symmetry is derived based on characteristics of text components. With the help of Sobel edge of the enhanced text image, we propose region growing to segment the complete using text representatives. In this way, the proposed method is different from the literature and is effective for video text segmentation.

## 2. Proposed Methodology

We observe that the color values of text pixel in R, G and B sub-bands usually are high values compared to its background values because the fact that text pixel have high contrast value compared to non-text pixel. To extract such observation, we propose Max-Min clustering on three values in R, G and B for each pixel to identify high contrast value to replace the pixel value in the image. As a result, we get high values for text pixels and low values for non-text pixels which is called enhanced image. Further to increase the gap between text and non-text pixels, we again propose Max-Min cluster with same criteria to identify the high contrast values from the neighbors. This results in sharpen image where text pixel are brighter than the pixel in the enhanced image. Since sharpened image increases the gap between the text and non-text pixels, we apply k-means with k=2 to obtain text cluster. To analyze the components in the text cluster, the method obtains skeleton image and the skeletons are checked whether they are fully connected components or not. We believe that at least one of the text components in skeleton image satisfies the fully connected component condition. This operation eliminates most of the background information as they do not satisfy the fully connected component criteria. We call them as text candidates. Due to complex background, there are chances that non-text components are considered as text candidates. Therefore, we propose novel symmetry criterion which is computed based stroke width of the text components. The stroke width is estimated by analyzing the gradient direction of each pixel of text components. The output of this operation is called as text representative image. For each text representative, the method applies region growing to grow the contour of the text representative along text direction in Sobel edge image of the enhanced image to segment the complete text. The flow diagram can be seen in Figure 1 where it shows all the steps of the proposed method.

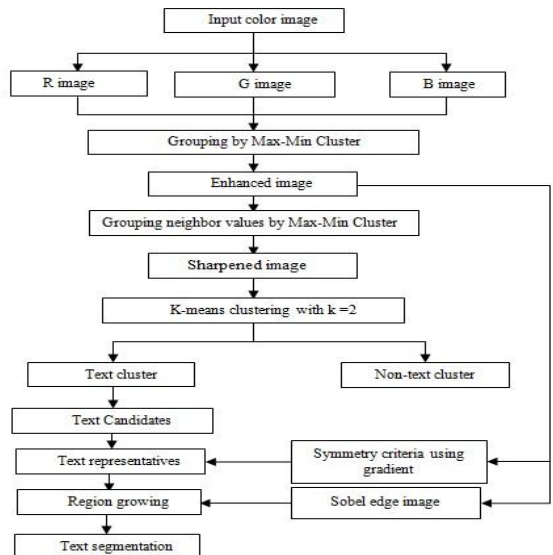

Figure 1. Flow diagram of the proposed method

## 2.1 Grouping Color Values for Text Enhancement

For the color input image shown in Figure 2(a), the method obtains three color sub-band images that are R-image, G-image and B-image shown respectively in Figure 2(b)-(d). In order to identify the high intensity value in three sub-bands, we propose Max-Min clustering criteria which select Maximum (Max) and Minimum (Min) value from R, G, B sub-bands for each pixel. Then the third value is compared with Max and Min values to find its closest value. If the third value is close to Max value then it forms a Max cluster with Max value. The method selects maximum value in the Max cluster to replace actual pixel value. Similarly, the actual pixel will be replaced by the minimum value if the third value is close to Min value and it forms a Min cluster. In this way, the Max-Min cluster does grouping to identify the high intensity values for each pixel in the input image which results in enhanced image as shown in Figure 2(e) where one can see the text pixels are brightened compare to the pixels in three sub-bands and input image. The basis for this is that usually text pixel has high intensity value in any one of three sub-bands compared to its background.

**Illustration of entire procedure using 4-by-4 image:**

Input:
```
R:  73  64  65  60      G:  79  70  69  66      B:  85  76  83  81
    75  74  78  86          81  79  86  93          89  82  87  97
    58  48  47  45          65  55  57  55          75  61  66  64
    43  71  48  40          49  78  56  50          63  86  67  63
```

Maximum values:        Minimum values:        Third values:
```
 85  76  83  81         73  64  65  60         79  70  69  66
 89  82  87  97         75  74  78  86         81  79  86  93
 75  61  66  64         58  48  47  45         65  55  57  55
 63  86  67  63         43  71  48  40         49  78  56  50
```

Clusters formed upon comparisons:

Max cluster:                                Min cluster:
{85, 79}  {76, 70}  {83}      {81}          {73}      {64}      {69, 65}  {66, 60}
{89}      {82, 79}  {87, 86}  {97, 93}      {81, 75}  {74}      {78}      {86}
{75}      {61, 55}  {66, 57}  {64, 55}      {65, 58}  {48}      {47}      {45}
{63}      {86}      {67}      {63}          {49, 43}  {78, 71}  {48, 56}  {50, 40}

After max-cluster and min-cluster method
```
 85   76   65   60
 75   82   87   97
 58   61   66   64
 43   71   48   40
```

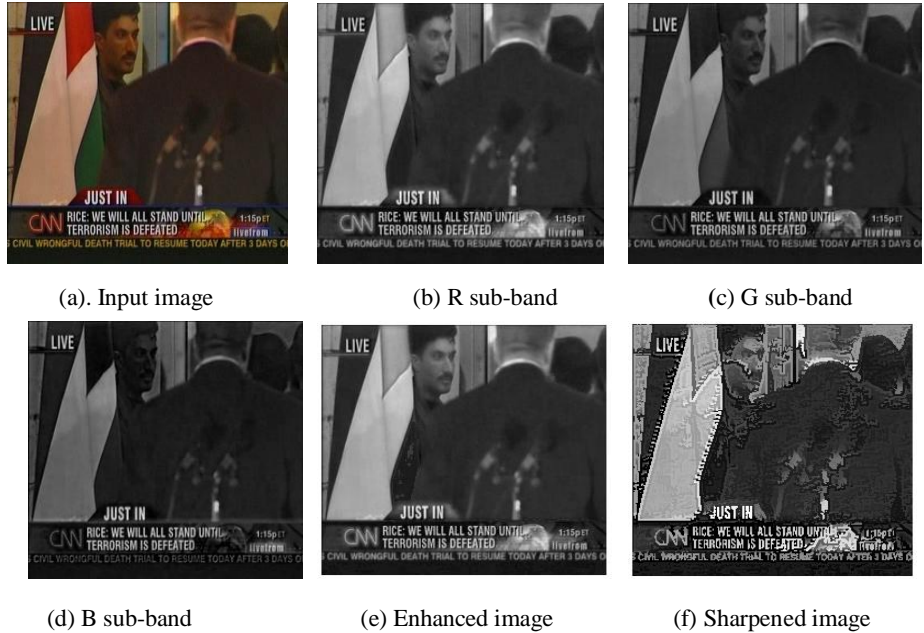

(a). Input image      (b) R sub-band      (c) G sub-band

(d) B sub-band      (e) Enhanced image      (f) Sharpened image

Figure 2. Steps to widen the gap between text and non-text pixels

## 2.2 Grouping Neighbor Values for Sharpening Text

It is true that text pixel must have high value compared to its neighbors because of high contrast of text pixels. To increase the gap between text and non-text pixels, we propose sliding window operation where we use the above process to sharpen the text pixel and to suppress the non-text pixel based on neighbor information. As a result, we get sharpened image as shown in Figure 2(f) where the text pixel are still brighter than the pixel in the enhanced image shown in Figure 2(e).

**Illustration of entire procedure using 4-by-4 image with two iterations:**

Input: Modified gray image:

```
85   76   65   60
75   82   87   97
58   61   66   64
43   71   48   40
```

Max = 87   Min = 58

Max cluster:                                       Min cluster:
{87, 85}   {87, 76}   {65}                         {85}       {76}       {65, 58}
{87, 75}   {87, 82}   {87, 87}                     {75}       {82}       {87}
 {58}       {61}       {66}                        {58, 58}   {61, 58}   {66, 58}

Updated matrix after first iteration and input for next iteration:

```
87   87   58   60
87   87   87   97
58   58   58   64
43   71   48   40
```

Max=97  Min=58

Max cluster:                                      Min cluster:
{97, 87}   {58}     {60}              {87}      {58, 58}   {60, 58}
{97, 87}   {97, 87} {97, 97}          {87}      {87}       {97}
 {58}      {58}     {64}              {58, 58}  {58, 58}   {64, 58}

Updated matrix after iteration2 and input for next iteration:

```
87   97   58   58
87   97   97   97
58   58   58   58
43   71   48   40
```

## 2.3 Text Candidates

The above two methods presented in Sections 2.2 and 2.3 help in widening gap between text and non-text pixels. This clue inspired us to use k-means clustering algorithm with k = 2 on sharpened image shown in Figure 2(f) to separate the text cluster. Since k-means clustering is unsupervised method, we consider cluster that gives high mean value compared to other mean of cluster as a text cluster. It is valid because high values will be classified into one cluster and that must be text cluster. The output of k-means can be seen in Figure 3(a) where almost all text pixel are classified as text including few high contrast non-text pixel.

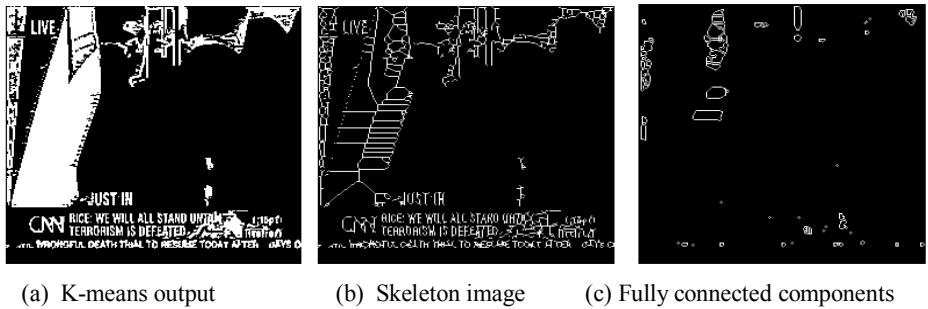

(a) K-means output        (b) Skeleton image        (c) Fully connected components

Figure 3. Steps to obtain text candidates

As it is noted that text cluster given by k-means provides binary information of both text and non-text components as shown in Figure 3(a). To reduce the pixel width to single pixel and preserve the shape of the text component, we get skeleton of the components using skeleton method as shown in Figure 3(b) where due to low resolution and complex background, skeleton may not preserve the shape of the components but it gives significant information to study the characteristics of text and non-text. For the purpose of removing non-text components from the skeleton image, we test whether text components satisfies fully connected component condition or not because it is fact that text components must be connected without any disjoint compared to non-text components where most of them are disjoint. The output of performing fully connected component testing is shown in Figure 3(c) where one can see most of the non-text components are removed and we call this output as text candidate image. The fully connected component is defined as starting and ending of the contour of the component should meet at one point. However, still we can see some non-text components in the results shown in Figure 3(c). We propose novel feature for text component verification in the following section.

## 2.4 Symmetry Criteria for Text Representative

We are inspired by the work presented in [10] where the stroke width concept is used for text detection and text component separation successfully. It works based on the fact that the stroke width is constant throughout the character while for non-text the stroke width distance is arbitrary. For each pixel in the fully connected component, the method computes stroke width distance that is traversing in perpendicular direction to the gradient direction of the pixel till it reaches white pixel which we call reached pixels. In this way, the stroke width is computed for each pixel in the fully connected component. Then to find dominant stroke width distance, we plot as histogram for the stroke width distances to choose the stroke width distance which gives highest peak as a dominant stroke width distance. Due to low resolution and complex background, it is hard to get complete shape of the characters. Therefore, one cannot expect constant stroke width for the character and hence we choose dominant stroke width distance for verification. For each reached pixel of the component, we obtain dominant stroke width distance. Then the method compares those two dominant stroke width distances to test the symmetry criteria. If both the distances are same then it is considered as the text component as it satisfies the symmetry criteria. This is true because according to stroke width concept presented in [10], the stroke width distance of the starting pixel and the reached pixel should be same. Note that the gradient image is obtained by performing the vertical and horizontal mask operation on the enhanced image as shown in Figure 4(a) and Figure 4(b) and the combined gradient image is shown in Figure 4(c) to estimate stroke width distance for the pixels in the fully connected components. The result of symmetry verification is shown in Figure 4(d) where almost all non-text components are removed and these components are called as text representatives.

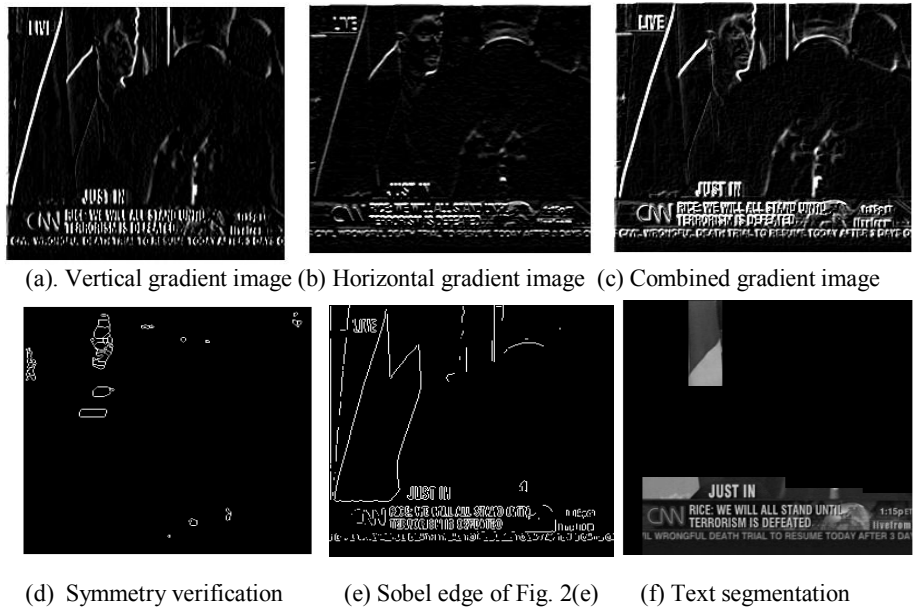

(a). Vertical gradient image (b) Horizontal gradient image (c) Combined gradient image

(d) Symmetry verification (e) Sobel edge of Fig. 2(e) (f) Text segmentation

Figure 4. Steps for text segmentation

## 2.5 Region Growing For Text Segmentation

The text representatives obtained by the previous section are considered as seed points to segment full text in the image. Therefore, we propose region growing method using Sobel edge map of the enhanced image as shown in Figure 4(e) for the purpose of text segmentation. We believe that the symmetry verification gives at least one seed point for one text line. The contour of the seed point grows pixel by pixel till it reaches white pixel of the neighbor component along text direction in Sobel edge map of the enhanced image. This process continues till end of the text line. The end of the text line is determined based on experimental study on space between the words and characters. The region growing works based on the assumption that the space between text lines is greater than the space between the words and characters. The main advantage of the region growing is that it segments text line of any orientation because it works based on nearest neighbor concept. For example, nearest neighbor for first character in the text line would be second character and for second character, third character is the nearest neighbor. Therefore, generally the region growing segment text individual text lines but sometimes due to noisy pixel between the text lines and disconnections, the method combines two to three adjacent text lines as shown in Figure 4(f) where the method segment four text line as one segment. This is the main drawback of this method. As a result, there is a scope for improving the region growing method so that it segments text lines properly.

## 3. Experimental Results

The proposed method is tested on variety of video frames to evaluate the performance of the method in terms of recall, precision and f-measure. Here we consider dataset provided by National University of Singapore (NUS) which contains video frames, extracted from news programmers, sports videos and movie clips. In this dataset, there are both graphic text and scene text of different languages, e.g. English, Chinese and Korean [20]. We consider 50 video text images to determine the recall, precision and f-measure in this work. Since we use small dataset to test the effectiveness of the proposed method, we test our method on large data by comparing with the existing methods in future.

To judge the correctness of the text blocks detected, we manually count the Actual Text Blocks (ATB) in the frames in the dataset and are considered as a ground truth. Since the main objective of the method is to segment text line in video frame, we consider each segmentation result given by the method as text block if it contains full text information else non-text block. Note that it is not in the line of text line detection in video. Based on this view, we define the following.

Truly Detected text Block (TDB): a detected block that contains text fully. Falsely Detected text Block (FDB): a detected block that does not contain text. The recall, precision and f-measures are defined as the following.

Recall (R) = TDB/ATB, Precision (P) = TDB/(TDB+FDB) and f-measure = 2RP/(R+P)

Sample results of the proposed method is shown in Figure 5 where first column and third column shows input images, and the second and the fourth column shows respective output images given the proposed method. It is observed from the results shown in Figure 5 that the proposed method works well for low resolution, complex background, different fonts and font sized images. For the image in third row and first column, and third row and third column in Figure 5, the proposed method segments text properly though the images contain complex background and small fonts. This is the advantage of the proposed method compared to the existing methods [20-21] where generally the methods fail to segment low contrast, low resolution and small font text.

The quantative results of the proposed method is reported in Table 1 where it is noticed that the proposed method gives promising and encouraging results as f-measure is more than 80% accuracy.

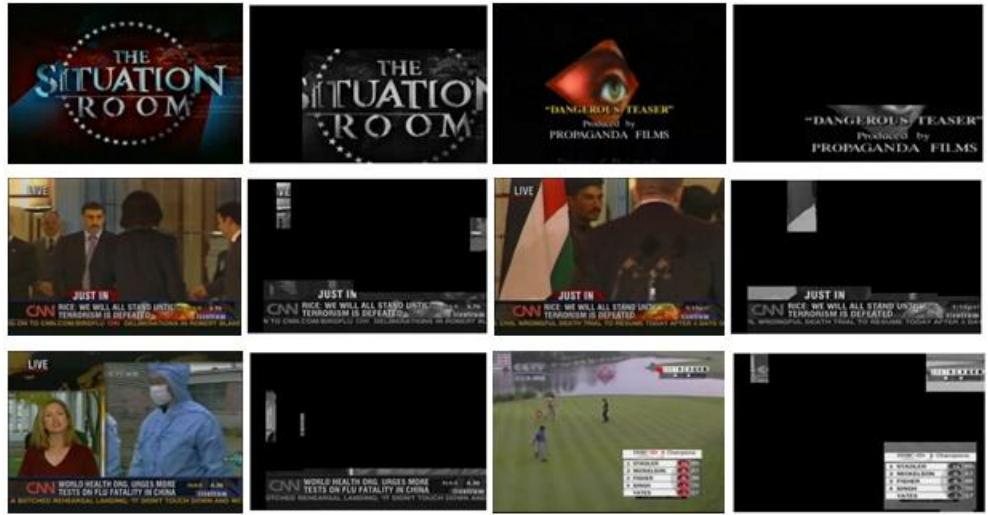

Figure 5. Sample results of the proposed method

Table 1. Quantative measures of the proposed method

| Recall | Precision | F-measure |
| --- | --- | --- |
| 0.85 | 0.84 | 0.82 |

## 4  Conclusion and Future Work

In this work, we have proposed new enhancement method by exploring color value in different sub-bands. We propose Max-Min clustering method to obtain Max and Min clusters to identify the high color value of the pixel. We also use neighbor information to enhance the enhanced image further to increase the gap between text and non-text pixel based on sliding window operation. We propose new criteria that checks whether the components obtained by the k-means clustering and skeleton are fully connected component are not. This step helps in eliminating most of the non-text components. Further, we introduce new symmetry verification based on stroke width distance of the text components to eliminate the non-text components and to obtain seed points for each text line. The region growing method is proposed to segment text lines by referring Sobel edge map of the enhanced image. The experimental result shows the proposed method is good for text segmentation from video frames. It is observed from the experimental results that the method gives low accuracy because the region growing merges two lines as one line while growing. We are planning to modify the region growing to segment text lines of any orientation in future and to improve the accuracy. The robustness of the method for handling complex backgrounds, different arbitrary orientation will be further investigation as well.

# References

1. N. Sharma, U. Pal and M. Blumenstein, "Recent Advances in Video Based Document Processing: A Review," In Proc. DAS, 2012. 63-68.

2. J. Zang and R. Kasturi, "Extraction of Text Objects in Video Documents: Recent Progress", In Proc. DAS, 2008, pp 5-17.

3. K. Jung, K.I. Kim and A.K. Jain, "Text information extraction in images and video: a survey", Pattern Recognition, 2004, pp. 977-997.

4. D. Doermann, J. Liang and H. Li, "Progress in Camera-Based Document Image Analysis", In Proc. ICDAR, 2003, pp. 606-616.

5. K. Jung, "Neural network-based text location in color images", Pattern Recognition Letters, 2001, pp. 1503-1515.

6. Q. Ye, Q. Huang, W. Gao and D. Zhao, "Fast and robust text detection in images and videos frames", Image and Vision Computing, 2005, pp 565-576.

7. D. Chen, J. M. Odobez and H. Bourlard, "Text detection and recognition in images and video frames", Patter Recognition, 2004, pp. 595-608.

8. L. Neumann and J. Matas, "Real-Time Scene Text Localization and Recognition", In Proc. CVPR, 2012, pp 3538-3545.

9. C. Yao, X. Bai, W. Liu, Y. Ma and Z. Tu, "Detecting Texts of Arbitrary Orientations in Natural Images", In Proc. CVPR, 2012, pp 1083-1090.

10. B. Epshtein  E. Ofek and Y. Wexler, "Detecting Text in Natural Scenes with Stroke Width Transform", In Proc. CVPR, 2010, pp. 2963-2970.

11. A. K. Jain and B. Yu, "Automatic Text Location in Images and Video Frames", Pattern Recognition, 1998, pp 2055-2076.

12. V.Y. Mariano and R. Kasturi, "Locating Uniform-Colored Text in Video Frames", In Proc. ICPR, 2000, pp 539-542.

13. V. Wu, R. Manmatha and E. M. Riseman, "Text finder: An automatic system to detect and recognize text in images", IEEE Transactions on PAMI, 1999, pp 1224-1229

14. K. L Kim, K. Jung and J. H. Kim "Texture-Based Approach for Text Detection in Images using Support Vector Machines and Continuous Adaptive Mean Shift Algorithm", IEEE Trans. on PAMI, 2003, pp 1631-1639.

15. P. Shivakumara, T, Q. Phan and C. L. Tan, "A Laplacian Approach to Multi-Oriented Text Detection in Video", IEEE Trans. PAMI, 2011, pp 412-419.

16. P. Shivakumara, R. P. Sreedhar, T. Q. Phan, S. Lu and C. L. Tan, "Multi-Oriented Video Scene Text Detection through Bayesian Classification and Boundary Growing", IEEE Trans. CSVT, 2012, pp 1227-1235.

17. J. Zhou, L. Xu, B. Xiao and R. Dai, "A Robust System for Text Extraction in Video", In Proc. ICMV, 2007, pp 119-124.

18. C. Lu, C. Wang and R. Dai, Text Detection in Images Based on Unsupervised Classification of Edge-based Features", In Proc. ICDAR, 2005, pp 610-614.

19. E. K. Wong and M. Chen, "A new robust algorithm for video text extraction", Pattern Recognition, 2003, pp 1397-1406

20. D. S. Guru, S. Manjunath, P. Shivakumara and C. L. Tan, "An Eigen Value Based Approach for Text Detection in Video", In Proc. DAS, 2010, pp 501-506.

21. M. Basavanna, P. Shivakumara, S. K. Srivatsa, and G. Hemantha Kumar, "A New Run-length based Method for Scene Text Detection", In Proc. IICAI, 2011, pp 1730-1736.